\pdfoutput=1

\documentclass[11pt]{article}

\usepackage{times}
\usepackage{latexsym}
\usepackage{amssymb}
\usepackage[colorinlistoftodos]{todonotes}
\usepackage[T1]{fontenc}
\usepackage{anyfontsize}
\usepackage{multirow}
\usepackage[final]{acl}

\usepackage[utf8]{inputenc}

\usepackage{microtype}
\usepackage{graphicx}
\graphicspath{{.}}
\usepackage{amsmath}

\title{Diverse Multi-Answer Retrieval with Determinantal Point Processes}

\author{Poojitha Nandigam$^*$, Nikhil Rayaprolu$^*$,  Manish Shrivastava \\
        Language Technologies Research Centre (LTRC) \\ International Institute of Information Technology, Hyderabad, India \\
        poojitha.nandigam@research.iiit.ac.in, nikhil.rayaprolu@students.iiit.ac.in, \\m.shrivastava@iiit.ac.in}
        
\begin{document}
\maketitle
\begin{abstract}

Often questions provided to open-domain question answering systems are ambiguous. Traditional QA systems that provide a single answer are incapable of answering ambiguous questions since the question may be interpreted in several ways and may have multiple distinct answers. In this paper, we address multi-answer retrieval which entails retrieving passages that can capture majority of the diverse answers to the question.  We propose a re-ranking based approach using Determinantal point processes utilizing BERT as kernels. Our method jointly considers query-passage relevance and passage-passage correlation to retrieve passages that are both query-relevant and diverse. Results demonstrate that our re-ranking technique outperforms state-of-the-art method on the AmbigQA dataset.
\def\thefootnote{*}\footnotetext{Equal contribution}\def\thefootnote{\arabic{footnote}}

\end{abstract}

\section{Introduction} 

The objective of open-domain question answering is to provide answers to queries utilising a large collection of documents from the World Wide Web, Wikipedia etc. 
More than 50\% of questions present in a widely used open-domain QA dataset (Natural Questions \citet{47761}) comprise of ambiguous questions (\citet{Min2020}).  
Questions that are ambiguous may be interpreted in a number of ways and as a result, they need various answers. In this paper, we concentrate on questions with multiple distinct answers.

Open domain question-answering systems are designed to generate answers from several data sources. Since similar information can be present across multiple data sources, it introduces a significant amount of redundancy. Traditional open-domain QA (\citet{Chen2017}) systems comprise of a Retriever, which retrieves passages relevant to the question. A passage retriever is primarily concerned with retrieving passages that are relevant to the query, and it does not address redundancy in the passages during retrieval. To be able to produce diverse answers to the question, the passages retrieved must be both relevant to the question and distinct from one another. After the retrieval stage, we introduce a novel re-ranking approach to handle redundant passages. As a result, the re-ranked passages would capture most of the diverse answers to the question. In this paper, we investigate the multi-answer retrieval task, which entails retrieving passages that can cover the distinct answers.
Re-ranking methods have been employed previously to improve the question answering accuracy significantly(\citet{wang2019multi};\citet{nogueira2019passage}; \citet{min2021joint}; \citet{clark2017simple}). 
\citet{min2021joint} tackles diverse multi-answer retrieval by proposing a re-ranker based on an auto-regressive framework in which each passage selected is dependent on the passages chosen at a previous time step.

Determinantal Point Processes (DPP) (\citet{Kulesza2012}) are probabilistic models that are effective at identifying diverse subsets of elements from a collection while preserving quality. DPP methods have proven effective in natural language processing tasks where there is a need for diverseness. \citet{Cho2019}, \citet{Li2019}, and \citet{Cho2020},  \citet{sharghi2017query} have used DPPs to perform summarization by choosing salient but also diverse items to be included in the summaries. In this paper, we propose an unsupervised re-ranking technique for multi-answer retrieval utilising Determinantal point processes and BERT to model the kernels.


Our contributions can be summarized as follows: 
\noindent 1) We propose a re-ranking method based on determinantal point processes that focuses on diverse passage retrieval. 

\noindent 2) Since our approach is unsupervised, our method does not require a large amount of data unlike prior re-ranking methods\cite{min2021joint}. Instead, we rely on DPP to identify the most relevant passages to the question that are distinct from one another. 

\noindent 3) We demonstrate that our technique outperforms the state-of-the-art method on the AmbigQA dataset using $\mathrm{MRECALL}$ @ $k$ metrics.

\section{Related Work}

Many open domain question answering systems(\citet{Chen2017}; \citet{Yang2019}; \citet{Izacard2021}; \citet{Guu2020}; \citet{lee2019latent}) adopt the \textit{retriever-reader} method by retrieving the relevant documents and later applying neural techniques to predict the answer. The \textit{retriever-reader} method was first proposed by \citet{Chen2017}. DrQA(\citet{Chen2017}) uses Wikipedia as knowledge source and employs a sparse retrieval method using TF-IDF and a recurrent neural network to identify the answer spans. While \citet{Yang2019} adopts Anserini retriever(\citet{yang2017anserini}) using BM25 as the ranking function and BERT model (\citet{devlin2018bert}) as the reader. Sparse retrieval based methods, such as TF-IDF and BM25, face challenges when retrieving relevant passages that do not match the question's exact terms.
Dense retrieval-based approaches, on the other hand, overcome this problem by mapping each word into a vector space in which words with similar meanings tend to be closer together.
ORQA (\citet{lee2019latent}) and DPR (\citet{Karpukhin2020}) employ a question and passage encoder based on BERT and compute a relevance score.
Using this relevance score, the retriever retrieves the most relevant documents from the corpus. 

\section{Determinantal Point Processes for Re-ranking} \label{dpp}
\begin{figure}[thp]
  \includegraphics[width=\columnwidth]{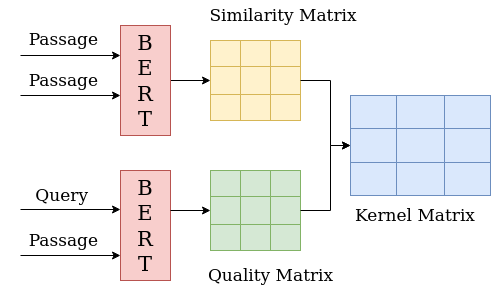}
  \caption{An overview of the proposed re-ranking method using DPP. A similarity score between the passages and a quality score between the question and passage are computed. These two scores are utilised to construct the DPP kernel matrix. }
  \label{fig1}
\end{figure}

Re-ranker acts as a filter to pick a limited number of passages that can be used as input to generate answers to the questions. We formulate the task of passage re-ranking as a subset selection problem.  Our objective is to choose a subset of passages ($Y$) of size $k$ from the ground set $\mathcal{Y}$ comprising $\mathsf{N}$ passages that covers all of the answers to a given question $q$ . DPP models a distribution on all the subsets of the ground set $\mathcal{Y}$ jointly considering the quality and diversity. A subset $Y$ is drawn according to the probability distribution $P$.
\begin{equation}
    P(Y;L) \varpropto det(L_Y)
\end{equation}
\begin{equation}  \label{eq2}
    P(Y;L)  = \frac{\det(L_Y)}{\det(L + I)}
\end{equation}
where $I$ is the identity matrix,  $L \in \mathbb{R}^{\mathsf{N}\times\mathsf{N}}$ is a positive semi-definite matrix referred as $L$-\textit{ensemble}, $\det(.)$ denotes the determinant of a matrix,  and $L_y$ is the submatrix of $L$  indexed by items in $Y$. $L$ matrix jointly considers query-passage relevance as well as passage-passage correlation through eq. \ref{eq3}.
\begin{equation} \label{eq3}
L_{ij} = Q(i, q)  \cdot S(i, j) \cdot Q(j, q)
\end{equation}
DPP focuses on two measures - quality and similarity ( Fig \ref{fig1}). Quality score $Q(i, q)$ measures how salient the passage $i$ is and whether it contains an answer to the question $q$. Similarity score $S(i, j)$ is computed between two passages $i$ and $j$ to incorporate diversity in the passages. DPP assigns a probability to a set Y proportional to the determinant of $L$-\textit{ensemble} which may be interpreted geometrically as the volume of the parallepiped covered by the quality and similarity measures (\citet{Kulesza2012}). A diverse passage subset occupies more volume than a subset of similar passages, therefore DPP assigns higher probability to diverse and relevant passages rather than the most relevant and similar passages.
If passages are relevant and diverse, then the passages can cover multiple distinct answers to the question.





\subsection{BERT for Similarity matrix}

To compute the similarity scores, we use a pretrained BERT model (\citet{devlin2018bert}; \citet{reimers-2019-sentence-bert}) to generate embeddings for every passage. The model takes the passage as input and produces a 768 dimensional dense embedding. We use these embeddings to calculate the cosine similarity of all passages and compute a similarity matrix $S \in \mathbb{R}^{\mathsf{N}\times\mathsf{N}}$ for the whole passage set. All the values in the similarity matrix lie in the range of $\lceil 0, 1 \rceil$. If passages $i$ and $j$ are similar, the similarity value $S(i, j)$ lies closer to $1$, if they are distinct, the value lies closer to $0$, and if $i = j$, $S(i,j)$ becomes equal to 1.
\begin{equation}
S(i, j) = cosine\_sim(BERT_A(i), BERT_A(j))
\end{equation}

\subsection{BERT for Quality matrix} \label{dppq}
We use a pretrained BERT model trained on MS MARCO (\citet{nguyen2016ms}) for computing the Quality matrix. The model takes in a query and a passage and generates the quality score. Higher quality score indicates that the passage is most relevant to the query and therefore most likely to answer the query. Unlike for computing similarity matrix, we do not perform cosine similarity over the model's outputs to produce a score, instead, we use a BERT encoder that concatenates both query and passage and generates a score.  The quality matrix $Q \in \mathbb{R}^{\mathsf{N}\times\mathsf{N}}$ is computed by performing the matrix multiplication of the scores ($N \times 1$) with it's transpose resulting in $N \times N$ dimensioned vector . These quality scores are then normalized to lie between $\lceil 0, 1 \rceil$.
\begin{equation}
Q(i, j) = Norm(BERT_B([i;j]))
\end{equation}
\subsection{Sampling}
Traditional DPP sampling algorithms have higher run-time complexity when $L$ matrix is large. We apply an efficient sampling technique - BFGMInference (\citet{Li2019};\citet{chen2018fast}). BFGMInference approximates a greedy approach to select a passage that maximizes the $det(L_Y)$ and adds it to the passage subset.
\begin{equation}
\begin{array}{c}
f(Y)=\log \operatorname{det}\left(L_{Y}\right) \\
k =\underset{i \in \mathcal{Y} \backslash Y}{\arg \max } f(Y \cup \{i\})-f(Y)
\end{array}
\end{equation}

\begin{table*}
\centering
\begin{tabular*}{0.75\textwidth}{@{\extracolsep{\fill} } lcc}
\hline
\multicolumn{1}{c}{\multirow{2}{4em}{Models}} & \multicolumn{1}{c}{Top5} & \multicolumn{1}{c}{Top 10} \\
\cline{2-2} \cline {3-3}
\\ [-1em]
   & AmbigQA-Dev & AmbigQA-Dev \\
\hline
DPR$^{+}$(\citet{min2021joint})      & 55.2/36.3    & 59.3/39.6      \\
DPR$^{+}$ + \citet{nogueira2019passage}          & 63.4/43.1        & 65.8/46.4       \\
JPR(\citet{min2021joint})       & 64.8/45.2     & 67.1/48.2      \\
QRR       & 62.0/42.3     & 70.8/57.6      \\
DPP-R      & \textbf{66.9}/\textbf{53.5}     &        \textbf{72.8/58.8}\\
\hline
\end{tabular*}
\caption{Performance of various models on AmbigQA dataset. Each row contains the $\mathrm{MRECALL}$ @ $k$ metrics for single answer retrieval and multi-answer retrieval respectively. }
\label{table:1}
\end{table*}

\section{Experiments}
In this section, we discuss about the passage retrieval method, the dataset we used in our experiments, the evaluation metric, and the results of our experiments.
\subsection{Passage retrieval} \label{psgret}

Wikipedia is utilised as the corpus for retrieving passages for the questions. Each Wikipedia article is broken into multiple passages containing the same number of words. We retrieve query-relevant passages from Wikipedia using the Dense Passage Retriever (DPR) (\citet{Karpukhin2020}; \citet{lin2021pyserini}). DPR computes encodings for all passages extracted from the Wikipedia corpus and builds an index. The inner product of the query and passage encodings is used to determine the similarity scores between them. Passages with the highest scores are the ones that are most relevant to the query, and these passages are subsequently sent into the re-ranker as input.

\subsection{Dataset}
We evaluated our method on an open-domain question-answering dataset AmbigQA (\citet{Min2020}), which contains multiple-answer questions. The dataset was created from an anonymised collection of Google search queries submitted by users seeking information on different subjects. It consists of 14,042 question-answer pairs derived from the Natural questions dataset (\citet{47761}) and is split into train, validation, and test sets. Train set consists of 10,036 question-answer pairs, validation set consists of 2,002 examples, and test set consists of 4,042 examples.

\subsection{Evaluation metric} \label{metric}
 $\mathrm{MRECALL}$ @ $k$ (\citet{min2021joint}) is used to evaluate the re-ranking of passages for questions with diverse answers. As per this metric, if  a query has $n$ answers, the $k$ passages that are retrieved must cover all of the answers. If $n<=k$, all answers must be covered; if $n > k$, the passages retrieved must contain at least $k$ answers. A retrieval is deemed successful if the passages retrieved include all or at least $k$ of the answers to the query.
 

\subsection{Results}
We compare our technique to a few additional baselines, all of which were assessed using the $\mathrm{MRECALL}$ @ $k$ metric on the AmbigQA dataset.

\begin{itemize}
\item \textbf{DPR}$^{+}$ \citet{min2021joint} integrates REALM (\citet{Guu2020}) with DPR (\citet{Karpukhin2020}). As described in Section \ref{psgret}, DPR is a dense retrieval based technique that utilizes the FAISS library to retrieve the relevant documents. Encoders for the query and passage are initialized using REALM and the DPR training method is followed.

\item \textbf{DPR$^{+}$} + \citet{nogueira2019passage} employs DPR$^{+}$ for the first stage of retrieval and the re-ranking method in \citet{nogueira2019passage} is applied on the retrieved passages.

\item \textbf{JPR} \citet{min2021joint} employs DPR$^{+}$ as the initial ranker and an auto-regressive framework is adopted as a re-ranker to generate diverse passages.

\item \textbf{Query Relevance Re-ranking(QRR)}
In this method, we first calculate the quality scores for each passage (described in section \ref{dppq}) and then we sort the passages based on these scores to pick the top-$k$ passages. Here, similarity among the passages is not considered.


\item \textbf{DPP-R} We employ our method described in section \ref{dpp} to retrieve highly diverse and relevant passages.
\end{itemize}

We calculate the performance of diverse multi passage retrieval using the $\mathrm{MRECALL}$ @ $k$ measure described in section \ref{metric}. Evaluation on the AmbigQA dataset demonstrates that our approach outperforms existing re-ranking techniques. 
Our technique requires no human annotations for multi passage re-ranking while outperforming existing methods, as shown in Table \ref{table:1}. DPP is modelled to select a subset of high-quality and diverse passages, which contributes to the success of our method for this task. Experiments demonstrate that the DPP-based technique achieves promising results for retrieving passages containing diverse answers.


\section{Discussion}
\noindent \textbf{Impact on QA system's performance: } An Open domain question answering system's pipeline consists of three stages. 1) Retrieval 2) Re-ranking followed by 3) Answer extraction. Improvements in any of these stages significantly improve the overall system's ability to answer a question. \citet{nogueira2019passage}, \citet{min2021joint} have shown that the use of a re-ranker has led to end-to-end QA improvements. 
Based on the results presented in Table \ref{table:1}, the DPP method enhances re-ranking for both single and multi-answer questions. We believe that this improvement in re-ranking will also improve the overall performance of the end-to-end QA system. 

\noindent \textbf{Impact of diversity: } 
DPP-R and JPR retrieve diverse passages utilising DPP and auto-regressive framework, respectively. Other approaches like QRR, retrieve just passages that are relevant to the query and do not tackle passage redundancy. We observe that our approach using DPP performs better than the QRR method. In order to re-rank, QRR simply considers how relevant a passage is to the query, and it retrieves the top-k passages with the highest relevance score for a given query. On the other hand, DPP-R takes into account how relevant the passage is to the query and also how similar passages are to each other, in order to eliminate redundant passages leading to diversity in the retrieved passages. 
DPP-R and JPR outperform other methods that do not emphasise diversity in multi-answer retrieval. For single answer retrieval, DPP-R and JPR have fared better than other methods, with the minor exception that QRR beats JPR in top-10 re-ranking. This demonstrates that diversity is an important aspect to consider during the re-ranking stage. 

\section{Conclusion}
In this paper, we propose a DPP-based approach to improve the diverseness of the retrieved passages.  
 We compare our method to the state-of-the-art method and outperform it by $3\%$ (top $5$), $8\%$ (top $10$) for single-answer questions, and $18\%$ (top$5$)  and $21\%$ (top$10$) for multi-answer retrieval on AmbigQA dataset. 
\bibliography{anthology,custom}
\bibliographystyle{acl_natbib}





\end{document}